\documentclass[sn-mathphys-num]{sn-jnl}

\pdfoutput=1

\usepackage{amsthm}
\usepackage{xcolor}
\usepackage{pifont}
\usepackage{graphicx}
\usepackage{multirow}
\usepackage{mathrsfs}
\usepackage{textcomp}
\usepackage{manyfoot}
\usepackage{booktabs}
\usepackage{listings}
\usepackage{algorithm}
\usepackage{algorithmicx}
\usepackage{algpseudocode}
\usepackage[title]{appendix}
\usepackage{amsmath,amssymb,amsfonts}

\usepackage[acronym]{glossaries}
\usepackage{breakcites}
\glsdisablehyper

\usepackage[capitalize,noabbrev]{cleveref}

\definecolor{green}{RGB}{80,200,120}
\newcommand{\xmark}{\textcolor{red}{\ding{55}}}
\newcommand{\cmark}{\textcolor{green}{\ding{51}}}

\newacronym{dl}{DL}{Deep Learning}
\newacronym{ml}{ML}{Machine Learning}
\newacronym{gru}{GRU}{Gated Recurrent Unit}
\newacronym{ai}{AI}{Artificial Intelligence}
\newacronym{mlp}{MLP}{Multi-layer Perceptron}
\newacronym{lstm}{LSTM}{Long Short-Term Memory}
\newacronym{rnn}{RNN}{Recurrent Neural Network}
\newacronym{gcn}{GCN}{Graph Convolutional Network}
\newacronym{convnet}{ConvNet}{Convolutional Neural Network}
\newacronym{muquar}{MuQAR}{Multimodal Quasi-AutoRegression}
\newacronym{ddpm}{DDPM}{Denoising Diffusion Probabilistic Model}
\newacronym{nfppf}{NFPPF}{New Fashion Products Performance Forecasting}
\newacronym{gtm}{GTM-Transformer}{Google Trends Multimodal Transformer}
\newacronym{prisma}{PRISMA}{Preferred Reporting Items for Systematic Reviews and Meta-Analyses}

\begin{document}

\title{New Fashion Products Performance Forecasting: A Survey on Evolutions, Models and Emerging Trends}

\author{\fnm{Andrea} \sur{Avogaro}}
\author{\fnm{Luigi} \sur{Capogrosso}}
\author{\fnm{Andrea} \sur{Toaiari}}
\author{\fnm{Franco} \sur{Fummi}}
\author{\fnm{Marco} \sur{Cristani}}

\affil{\orgdiv{Dept. of Engineering for Innovation Medicine}, \orgname{University of Verona}}

\abstract{
The fast fashion industry's insatiable demand for new styles and rapid production cycles has led to a significant environmental burden.
Overproduction, excessive waste, and harmful chemicals have contributed to the negative environmental impact of the industry.
To mitigate these issues, a paradigm shift that prioritizes sustainability and efficiency is urgently needed.
Integrating learning-based predictive analytics into the fashion industry represents a significant opportunity to address environmental challenges and drive sustainable practices.
By forecasting fashion trends and optimizing production, brands can reduce their ecological footprint while remaining competitive in a rapidly changing market.
However, one of the key challenges in forecasting fashion sales is the dynamic nature of consumer preferences.
Fashion is acyclical, with trends constantly evolving and resurfacing.
In addition, cultural changes and unexpected events can disrupt established patterns.
This problem is also known as \gls{nfppf}, and it has recently gained more and more interest in the global research landscape.
Given its multidisciplinary nature, the field of \gls{nfppf} has been approached from many different angles.
This comprehensive survey wishes to provide an up-to-date overview that focuses on learning-based \gls{nfppf} strategies. 
The survey is based on the \gls{prisma} methodological flow, allowing for a systematic and complete literature review.
In particular, we propose the first taxonomy that covers the learning panorama for \gls{nfppf}, examining in detail the different methodologies used to increase the amount of multimodal information, as well as the state-of-the-art available datasets.
Finally, we discuss the challenges and future directions.}

\keywords{New Fashion Products Performance Forecasting, Computer Vision for Fashion, Multimodal Time Series Forecasting.}

\maketitle

\glsresetall

\section{Introduction} \label{cha:cha_introduction}

The challenge of predicting the sales performance of new fashion products, known as the \textbf{\gls{nfppf}} problem, is a critical issue in the apparel industry~\cite{cho2014learning,beheshti2014survey,singh2019fashion}, where companies must accurately forecast future sales of new products while having little or no historical data.
In an industry characterized by rapidly changing consumer trends and preferences, forecasting demand for new items is essential for optimizing production, inventory management, and marketing strategies.

Traditional time series forecasting methods, such as ARIMA~\cite{box2015time} or other statistical models, such as~\cite{hyndman2008forecasting,box2015time,taylor2018prophet}, heavily depend on historical data.
Similarly, newer \gls{dl} models~\cite{salinas2020deepar,oreshkin2020n} operate under the same assumption, which may prove inadequate when forecasting new products without prior sales history.
This creates a unique challenge with far-reaching implications for supply chain efficiency, financial planning, and sustainability.

Addressing this problem, first introduced by~\cite{ren2017comparative,arvan2019integrating,lara2021experimental}, is crucial due to several issues.
Firstly, inaccurate demand forecasting can lead to overproduction, resulting in overstocking, financial losses, and increased waste~\cite{fisher2001optimizing,taplin2014global}.
This issue is particularly severe in the fashion industry, which is under increasing scrutiny due to its environmental impact due to excess production and high return rates~\cite{pollution2022}.

Secondly, underestimating demand can result in inventory shortages, missed sales opportunities, and dissatisfied customers.
As a result, balancing supply and demand is essential, particularly in the face of volatile fashion trends and unpredictable consumer behavior~\cite{masyhuri2022key}.

\textbf{\emph{The problem of forecasting demand for new products.}}
The \gls{nfppf} task can have several different hues, trying to cover different real-world scenarios in the fast fashion industry. In this Section, we will try to define different use cases with different motivations, which can be grouped under \gls{nfppf}. 

The first scenario consists of forecasting the demand for new products, in which models need to estimate future sales of a product that has never been sold.
In this case, the challenge is to find patterns and similarities with older, more successful products while incorporating real-time market trends~\cite{singh2019fashion}.

When only a small number of initial observations are available, we can identify a second scenario of \gls{nfppf}~\cite{ekambaram2020attention,skenderi2024well}.
Here, the challenge is to make reliable forecasts with limited data, requiring models that can learn quickly from sparse sales data and adapt to changing trends.
These forecasts are critical to making timely decisions about replenishment or adapting marketing strategies after the item is released.

Finally, the third use case involves forecasting a product that has not yet been released and is still in the development phase, helping companies in the decision-making process~\cite{joppi2022pop}.
It involves building personalized models that understand the behavior, preferences, and purchase history of customers, allowing companies to suggest the most relevant products to each user.
Personalized recommendations increase customer satisfaction and stimulate sales by improving the relevance of offers.
This could be done to decide whether a product is worth producing or not, to propose garments only with a certain performance in the market.

In all three scenarios, the need for models that can effectively predict the amount of product that will be sold is shared.
However, the requirements with respect to the application case are different and satisfy different needs of clothing manufacturers.
Section~\ref{cha:cha_nfppf_taxonomy} lists several models that fall into these three specific cases also in relation to the datasets used in the experimental section.

\textbf{\emph{The role of learning-based models.}}
To effectively address the challenge of forecasting the performance of new products in the fashion industry, \gls{ml} and \gls{dl} models offer significant advantages over traditional methods.
Unlike classic time series forecasting models, which are limited in predicting new item sales with historical data, \gls{dl} models can leverage a range of product characteristics such as color, style, and category, but also with other marketing information such as discounts, promotions, and release dates.
In this way, they can emulate expert judgment, which has traditionally been used to predict new fashion items~\cite{hyndman2021forecasting}.
Through this emulation, these models improve the forecasting process by capturing similarities between new and existing products on the market.

\subsection{NFPPF: A Multimodal Problem}
\begin{figure*}[t!]
    \centering
    \includegraphics[width=\linewidth]{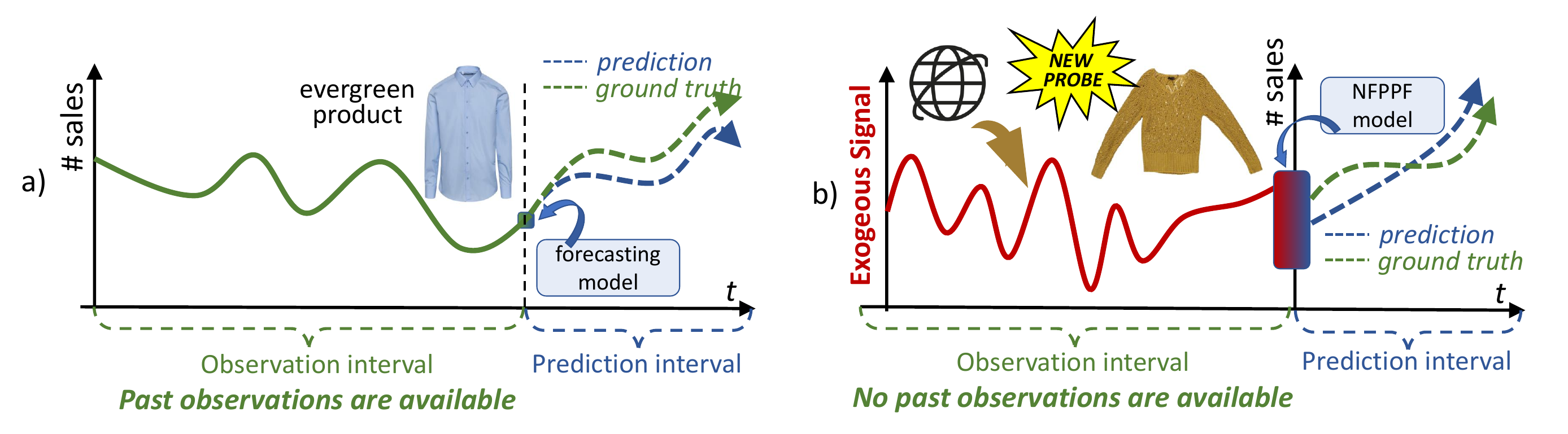}
    \caption{a) A standard forecasting setup, where an evergreen item has past observations to exploit, \emph{e.g.}, \# sales.
    b) \gls{nfppf} problem, where no past observations are available, and exogenous data must be considered. 
    The figure is from~\cite{joppi2022pop}.}
    \label{fig:fig_nfppf}
\end{figure*}

One of the major breakthroughs in this field has been the integration of visual data.
Although early models focused primarily on textual attributes~\cite{bahdanau2016neural,singh2019fashion,craparotta2019siamese}, such as product description, adding visual information, such as product images, has significantly improved accuracy.
Visual features often provide a richer and more nuanced understanding of products than textual descriptions alone, making image-based models more effective in predicting consumer demand~\cite{ekambaram2020attention}.
This shift reflects the realization that, in the fashion industry, appearance plays a key role in determining a product's success in the marketplace.

In addition to a simple analysis of product features, newer models have used external data sources to assess market trends and popularity.
For example, integrating real-time data streams, such as online search trends, allows forecasting models to better account for consumer interest as it evolves~\cite{skenderi2024well}.
This multisource approach provides a more holistic view of potential sales performance, as it captures both the intrinsic attributes of the product and the broader market context in which it is introduced.

Another advancement in this area involves the use of publicly available images that share some characteristics with the target product, giving a prior of its potential popularity.
By examining the performance of similar products on the market, these models can make more informed predictions, even without direct historical data on sales of the new product~\cite{joppi2022pop}.
This approach increases the predictive power of machine learning models by incorporating external knowledge and historical patterns to infer the likely outcomes of new releases.

Furthermore, the evolution of model architectures has also played a crucial role in improving forecasting results~\cite{capogrosso2024machine}.
Specifically, \gls{dl} models, such as those that use attention mechanisms, have enabled more granular and accurate forecasts by focusing on the most relevant features of each product~\cite{skenderi2024well}.
These models not only predict sales, but also provide insight into which product features spark consumer interest, giving brands useful feedback to refine their offerings.

In summary, integrating textual and visual data and taking into account real-time trends has proven to be very effective in improving the accuracy of forecasting demand for new products.
Companies can optimize decision-making processes at various stages by focusing on these multimodal inputs, including supply chain management and the development of marketing strategies.
As these technologies evolve, they promise to improve the fashion industry's ability to predict demand more effectively, ultimately fostering more sustainable practices by reducing overproduction and waste.

\subsection{NFPPF: Available Datasets}
\begin{figure*}[t!]
    \centering
    \includegraphics[width=\linewidth]{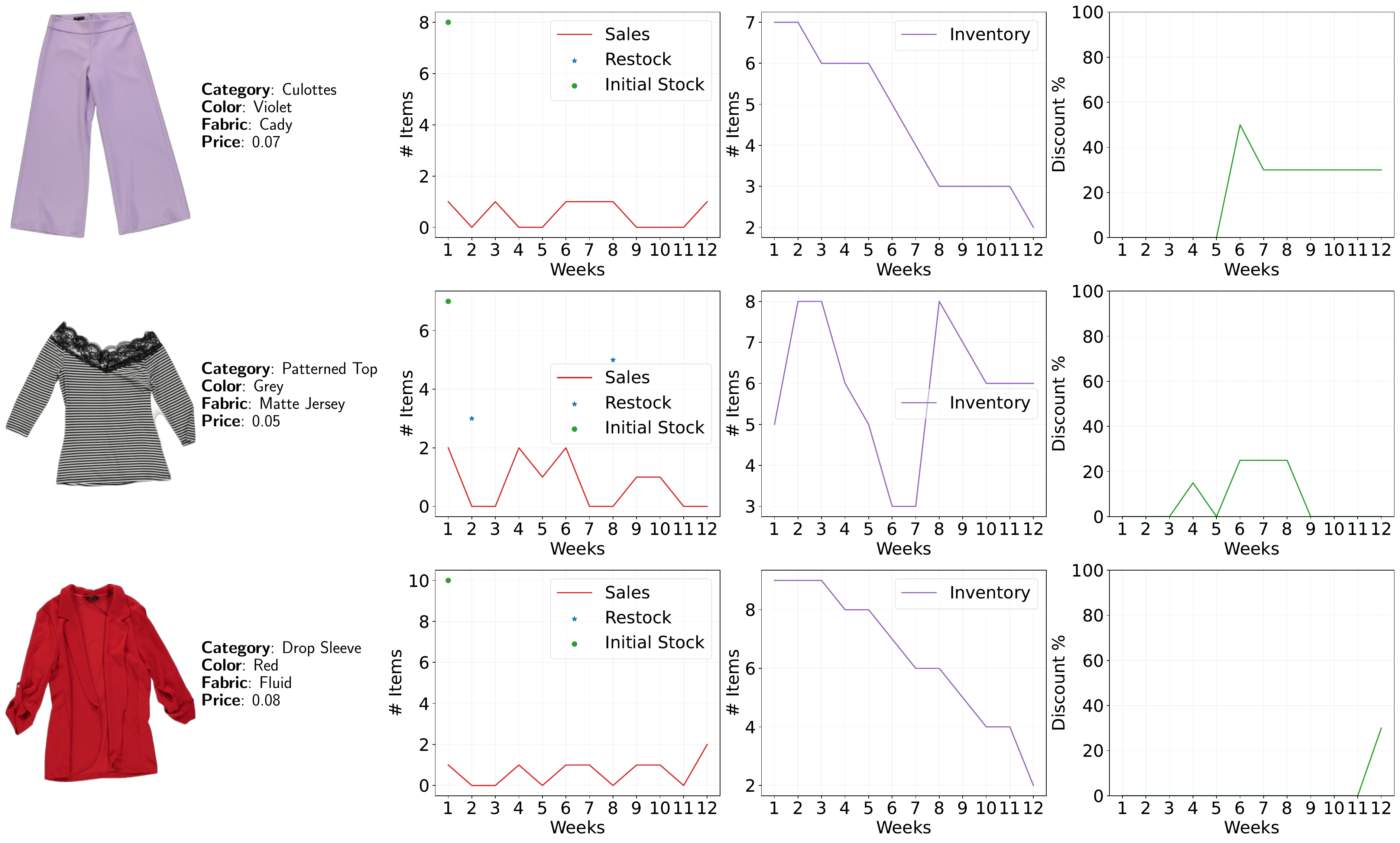}
    \caption{Examples of data in the VISUELLE 2.0 dataset.
    The figure reports (from left to right) the product’s image, textual tags, sales time series, restocking information, inventory time series, and discount time series.
    Figures are from~\cite{skenderi2022multi}.}
    \label{fig:fig_visuelle}
\end{figure*}

Publicly available datasets for fashion forecasting, like~\cite{hmdataset}, consider diverse applications different from \gls{nfppf}.
They have usually been used to forecast fashion trends, aggregating products of multiple brands and their popularity on social networks, \emph{e.g.}, Instagram.
In \gls{nfppf}, the task is different since the focus is only on single products and not on groups of products. 
So, there are less data with which to reason. 

In~\cite{skenderi2024well}, the authors propose the VISUELLE dataset, the first dataset in the literature for \gls{nfppf} and the de facto standard for this task.
In particular, VISUELLE contains sales data for 5,577 different products registered in 100 stores of the Italian company Nunalie, spanning the period between October 2016 and December 2019.
Each product is represented by an RGB image with resolutions that vary from 256 to 1,193 pixels (width) and from 256 to 1,172 pixels (height), and each image portrays the clothing item on a white background without the person wearing it.
In addition, a binary foreground mask is provided.
Each product is further annotated with several tags, validated by the Nunalie team, and includes category, fabric, and temporal information.
Additionally, the dataset provides weekly sales time series and Google Trends data to analyze the popularity of search terms associated with visual content.

In~\cite{skenderi2022multi}, this dataset was further expanded with a second version.
VISUELLE 2.0 describes sales between November 2016 and December 2019 of 5,355 different products in 110 different stores.
In addition to the information mentioned above, in this version of the dataset, the authors added anonymized data for 667,086 customers who have requested a fidelity card, allowing them to extract the history of their purchases and the baskets of products they bought.
These data consist of the ID of the purchased product, the date-time of purchase, the retail store ID, and the quantity.
Finally, weather reports downloaded from IlMeteo\footnote{https://www.ilmeteo.it/portale/archivio-meteo.} are also provided, which contain the daily weather conditions at the municipality level.

\subsection{Motivation and Contributions}
This survey aims to serve as a valuable resource for researchers and industry practitioners seeking to develop more sustainable, efficient, and accurate fashion forecasting models that align with the evolving landscape of the fashion industry.

Specifically, our contributions are as follows.
\begin{itemize}
\item We provide a detailed examination of the existing literature on learning-based models for \gls{nfppf}, highlighting their strengths, limitations, and unique approaches.
\item Analyzing current research, we discuss possible future directions, proposing a novel framework that integrates advanced predictive analysis, \emph{e.g.}, the use of human proxemics, to solve the forecasting problem in the fashion industry.
\end{itemize}

\subsection{Article Organization}
The survey is organized as follows.
\cref{cha:cha_selection_criteria} describes the article selection criteria for creating this systematic review.
\cref{cha:cha_nfppf_taxonomy} is the core of the survey and reviews the collection of algorithms and techniques for \gls{nfppf}.
\cref{cha:cha_trends} introduces our proposed frameworks and discusses potential directions for future research.
Finally,~\cref{cha:conlusion} concludes the survey.

\section{Selection Criteria} \label{cha:cha_selection_criteria}

\begin{figure*}[t!]
    \centering
    \includegraphics[width=\linewidth]{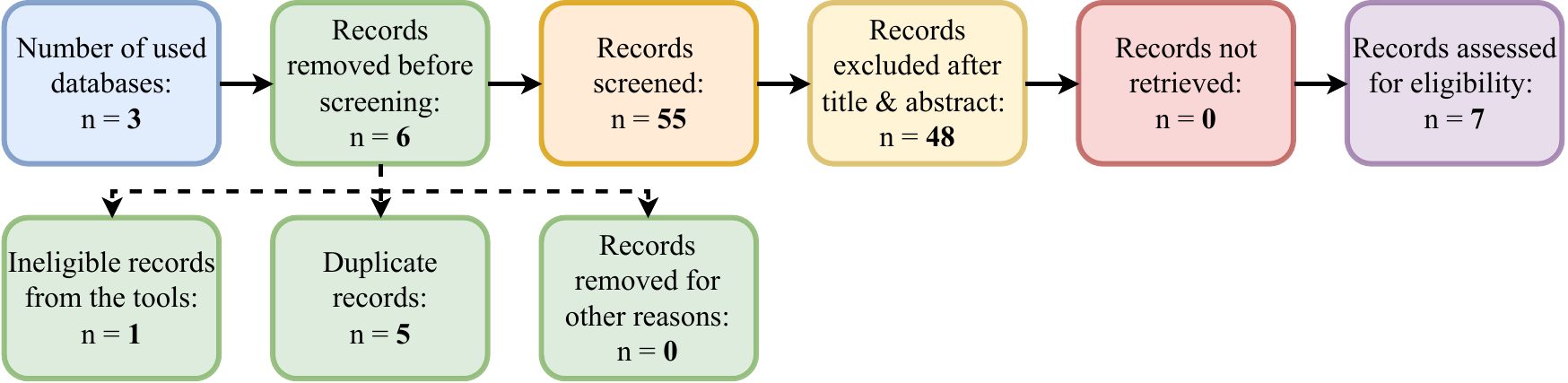}
    \caption{\gls{prisma}-based flowchart of the retrieval process. 
    Due to space issues, the diagram has been reformatted into a horizontal rather than the standard vertical format.}
    \label{fig:fig_prisma}
\end{figure*}

This section describes the selection criteria for this systematic review and how the articles were collected.

Only publications in English were considered, and all the studies had to be published in peer-reviewed journals or conference proceedings.
The search strategy and selection criteria were developed in consultation with all authors.
Any disagreements between authors were resolved through discussion and consensus.
To gather up-to-date knowledge from a broad spectrum of information sources, this survey was conducted following a widely known systematic literature review methodology known as \gls{prisma} guidelines, the golden standard for improving transparency, avoiding missing citations, and completeness in documented systematic reviews and meta-analyses.

The included studies were extracted from the following three databases: \textbf{\textit{Web of Science}}, \textbf{\textit{Scopus}}, \textbf{\textit{IEEE Xplore}}, covering the period between January 2020 and December 2024.
All search queries included the following terms: \textit{``New Fashion Products Performance Forecasting''}, \textit{``Fast Fashion''}, \textit{``Fashion Sales Forecasting''}, \textit{``VISUELLE''}, \textit{``VISUELLE 2.0''}.
These concepts form the basis for the inclusion criteria for selecting studies considered in the systematic review.
Therefore, cited papers in this work contain the above keyword combination.

Our keywords produced a total of 61 records.
\cref{fig:fig_prisma} illustrates the \gls{prisma} flow chart, a transparent and replicable means of reporting the systematic review search and selection process.
First, we remove all duplicate papers (5 excluded).
Next, we excluded all the papers marked as ineligible by the automation tool (1 excluded) and non-accessible papers (0 excluded).
After the title and abstract screening process, 48 articles were selected to be removed.
The number of records not found is 0.
As a result, 7 were eligible.

Finally, none of the 7 reviewed articles were comprehensive survey papers on \gls{nfppf} techniques.
As a result, we claim that this is the first systematic review that addresses these topics.

\section{NFPPF: Evolutions and Models} \label{cha:cha_nfppf_taxonomy}

\begin{figure*}[t!]
    \centering
    \includegraphics[width=\linewidth]{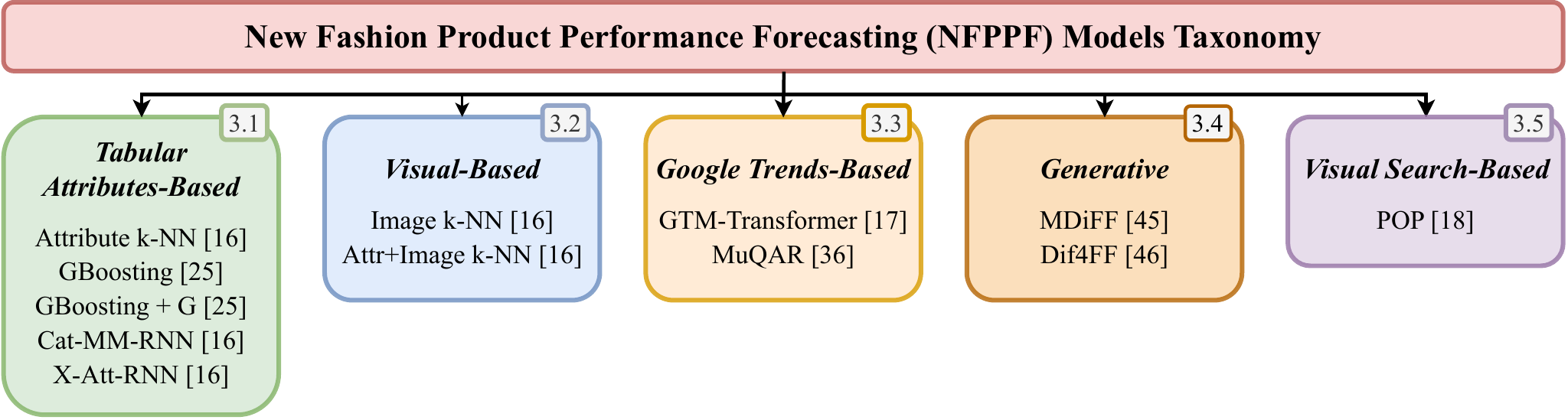}
    \caption{Our proposed taxonomy covers the available methodologies for addressing the \gls{nfppf} task.}
    \label{fig:fig_nfppf_taxonomy}
\end{figure*}

This section reviews the available methodologies for addressing the \gls{nfppf} task, dividing it into five subsections based on the main data modality used and the architecture.
Specifically, most of the methods share many of the discussed modalities, but we decided to group them by selecting the main modality or the one that the authors proposed as novel in that article.

We propose the taxonomy shown in~\cref{fig:fig_nfppf_taxonomy}, and the following sections will dive into the macro-areas of \textbf{tabular attributes-based models} (\cref{sec:sec_tabular_models}), \textbf{visual-based models} (\cref{sec:sec_visual_models}), \textbf{Google Trends-based models} (\cref{sec:sec_gtrends_models}), \textbf{generative models} (\cref{sec:sec_gen_models}), and \textbf{visual-search-based models} (\cref{sec:sec_visual_search_models}).
It is important to note that some models reported in the next sections are developed for datasets that provide historical data.
Those models are included for the sake of completeness of the literature, but are not strictly related to \gls{nfppf}.

\subsection{Tabular Attributes-Based Models} \label{sec:sec_tabular_models}
Early work in this domain has focused primarily on leveraging well-established \gls{ml} techniques, such as gradient boosting methods (\emph{e.g.}, XGBoost~\cite{friedman2001greedy}) and ensemble approaches such as Random Forest, to address this issue~\cite{loureiro2018exploring,singh2019fashion}.
These models have been fed with a wide range of static attributes, such as product category, color, and fabric, as well as dynamic information, such as promotions and discounts, which can significantly influence consumer behavior.

In the same paper~\cite{singh2019fashion}, the authors also explored the effectiveness of \gls{dl} architectures, including \glspl{mlp} and \gls{lstm} networks, which were used to process product attributes and generate more nuanced forecasts compared to the \gls{ml} algorithm previously explained.
Despite their sophisticated structure, these models were still limited, as they relied solely on structured tabular data, such as product attributes and promotional strategies.
Furthermore, while these models demonstrated some potential, they did not fully explore the integration of multimodal data in sales trends, which could provide a more comprehensive picture of how products are likely to behave under real-world market conditions.

More recent advancements~\cite{ekambaram2020attention} have started to address these gaps by incorporating additional data types and more complex neural network architectures.
For example, architectures based on \glspl{rnn} have shown promise in capturing sequential patterns in sales data, allowing the incorporation of static attributes and time-varying signals such as past sales performance, discount timelines, and consumer engagement trends.

\subsection{Visual-Based Models} \label{sec:sec_visual_models}
As explained in~\cref{cha:cha_introduction}, the most difficult part of predicting how much a garment will sell relies on the absence of past information or signals.
In most of the proposed methods presented in~\cref{sec:sec_tabular_models}, a large portion of the contributions rely upon how to process different types of tabular information related to the garment itself.
Despite their importance, what really drives people to buy or not buy a garment is something far from being quantitative at first sight.
Fashion trends are often completely new, unpredictable, and not expressible as tabular data.
Therefore, other viable sources of information are needed.

As a result, in~\cite{ekambaram2020attention}, the authors proposed, for the first time, the use of a \gls{convnet} to integrate visual attributes.

The inclusion of image features marks a significant shift in how models are designed, as it allows algorithms to capture visual similarities between new and existing products, which can be a critical factor in consumer decision-making processes.
By integrating this diverse range of static, dynamic, and unstructured signals, these newer \gls{ml} approaches are better equipped to forecast the sales performance of new products. 

Specifically, in~\cite{ekambaram2020attention}, the authors introduced a notable advancement by developing an autoregressive \gls{rnn}-based model designed for sales forecasting.
This model incorporates various inputs, including past sales data, auxiliary information such as release dates and discounts, textual embeddings of product descriptions, and, most importantly, product images.
To determine which input modalities contribute the most to sales, the model uses an additive attention mechanism~\cite{bahdanau2016neural}.
The attention mechanism highlights the most relevant features, which are then integrated into a feature vector processed by a \gls{gru} decoder~\cite{cho2014learning}.

The experimental results showed that this architecture based on \gls{rnn} excels in capturing expert judgments due to its ability to represent products through their visual, textual, and temporal attributes.

Furthermore, one of the model's key strengths is its interpretability, as the attention mechanism reveals which input modality is most influential at each step of the forecast.

However, this article does not address crucial factors such as the impact of trends and fashion popularity on predictions.

\subsection{Google Trends-Based Models} \label{sec:sec_gtrends_models}
The first model that used exogenous signals like Google Trends is \gls{gtm}~\cite{skenderi2024well}, a non-autoregressive Transformer architecture designed to predict the sales of new products by integrating information from several modalities.

Unlike traditional models~\cite{hyndman2021forecasting,lara2021experimental}, \gls{gtm} aims to address the challenge of forecasting new products by considering product characteristics and trends, originally introduced in~\cite{choi2014fast,beheshti2014survey}.
Previous studies have shown that Google Trends can be used to predict a variety of time series, such as real estate sales and inflation~\cite{wu2009future,guzman2011internet,hand2012searching,hamid2015forecasting,bangwayo2015can,bulut2018google}.

Specifically, \gls{gtm} leverages four key modalities:
\begin{itemize}
\item Product Image: Visual data representing the product.
\item Textual Descriptors: Category, color, and fabric information.
\item Temporal Information: The planned release date of the product.
\item Exogenous Popularity Information: Google Trends data on the popularity of textual descriptors to provide insight into what is currently trending in the market.
\end{itemize}
In detail, the \gls{gtm} model consists of two main components: a Transformer encoder~\cite{vaswani2017attention} and a custom Transformer decoder.

The encoder processes a multivariate time series formed by Google Trends, which are derived from the textual description of the products.
Specifically, it uses scaled dot-product self-attention, which helps the model to focus on the most relevant parts of the input time series for the task.
The model enforces locality by masking the input with a block-diagonal matrix~\cite{rae2020transformers}, ensuring that it focuses primarily on information from the same or nearby time periods.
The encoder output is a representation of the Google Trends time series, enriched with an understanding of which parts are most important for forecasting future product popularity.

The decoder in~\gls{gtm} differs from the traditional Transformer decoder proposed by~\cite{vaswani2017attention}.
Instead of utilizing a self-attention layer, which is standard in many Transformer architectures, this model skips self-attention because the input is not a sequence, but a single vector produced by the feature fusion network.
This input vector serves as the query in the multi-head cross-attention mechanism, while the encoded Google Trends representation is used as the key and value.
The decoder essentially tries to learn which parts of the exogenous popularity signal are most relevant to the multimodal product embedding to generate an accurate forecast.
The final output of the decoder is a compact representation of the product, encapsulating the four modalities (visual, textual, temporal, and exogenous).
Finally, this representation is passed through a fully connected layer, which projects it into the final sales forecast.

Another similar work has been proposed by~\cite{papadopoulos2022multimodal}, proposing~\gls{muquar}.
MuQAR employs a multimodal \gls{mlp} (the authors call it FusionMLP) to process the visual and textual features extracted from the dataset's product images.
The visual features are generated from a \gls{convnet}, pre-trained on the COCO dataset~\cite{lin2015microsoftcococommonobjects}.
Then, a hierarchical \gls{dl} network is specifically fine-tuned for fashion imagery, ensuring that the model captures relevant visual attributes.

A key component of MuQAR is image captioning via the OFA model~\cite{wang2022ofa}, which extracts descriptive information from garment images.
These captions provide detailed contextual descriptions, such as color and fabric placement, enriching the standard attribute detection methods.
OFA's captions, which are pre-processed and tokenized, enable the model to generate a richer textual feature vector for improved forecasting performance.

The time series forecasting component of \gls{muquar} is handled by the Quasi-AutoRegressive (QAR) module, which models both the time series of the target fashion attributes and the exogenous attributes.
For the VISUELLE dataset, the time series of attributes such as color, fabric, and other features are treated as the target series, while other fashion attributes serve as exogenous input. 
This structure allows the model to predict the product's future popularity by learning relationships between various fashion attributes over time.

MuQAR's modular architecture allows for experimentation with different time series forecasting models, such as \gls{lstm}~\cite{hochreiter1997long}, \gls{convnet}-\gls{lstm}~\cite{zhao2017convolutional}, and DA-RNN~\cite{qin2017dual}, to identify the optimal architecture for \gls{nfppf}.
Depending on the architecture, these models can process the fashion attributes of the garment and related exogenous time series in parallel or sequential stages.
Using both target and exogenous attribute series, the model can produce accurate trend popularity forecasts for new garments, even without historical data.

By integrating multiple modalities and focusing on current market trends, both \gls{gtm} and \gls{muquar} models mimic the decision-making process of human experts, who often rely on product characteristics and trending information to make forecasts~\cite{hyndman2021forecasting}.

\subsection{Generative Models} \label{sec:sec_gen_models}
Given the fast-paced trends of the fashion industry, the definition of what is fashionable and what is not changes rapidly (\emph{i.e.}, in years or even seasons), making it difficult to understand exactly what the market performance could be for a new item and what factors are worth considering.
To date, ``classical'' deterministic forecast models have shown quasi-satisfactory performance in certain scenarios but generally lead to unrealistic predictions due to the shift in input features' domain.

In data modeling, diffusion-based approaches, exemplified by \glspl{ddpm}~\cite{ho2020denoising}, have emerged as powerful tools.
Their ability to capture the underlying probabilistic structure of data has proven priceless in tasks ranging from visual content generation to time series forecasting~\cite{lin2023diffusion}.
In particular, in the context of fashion sales forecasting, where data features can fluctuate rapidly, \glspl{ddpm} show a remarkable ability to adapt and maintain accurate predictions, even when faced with patterns of new or unknown features.
Unlike traditional deterministic models that rely on direct mappings between input features and output predictions, \glspl{ddpm} exploit a diffusion process that gradually refines the noisy data, ensuring that the predictions remain grounded in the observed data distribution~\cite{sohl2015deep}.
This adaptive nature makes \glspl{ddpm} a compelling choice for forecasting challenges characterized by non-stationarity and changing trends.

As a result, in~\cite{avogaro2024mdiff,avogaro2024dif4ff}, the authors present the first two-step multimodal diffusion-based pipeline for \gls{nfppf}.
Firstly, they build and train a multimodal score-based diffusion model, \emph{i.e.}, a generalization of a \gls{ddpm}, to provide initial predictions and handle cases with features beyond the training distribution.
Secondly, they refine the diffusion model outputs using a lightweight \gls{mlp} in~\cite{avogaro2024mdiff} and a powerful \gls{gcn} in~\cite{avogaro2024dif4ff}, which should be regarded as the extension of~\cite{avogaro2024mdiff}.
Despite the effectiveness of diffusion models, their output is based on an aleatory sampling process and can strongly differ from that of other models that generate outcomes starting from the same conditioning.
For this reason, they generate multiple predictions for each object, exactly 50, to be precise, and then use them as inputs for the refinement model to have the final prediction.
This strategy ensures smoother in-distribution data as input for the refinement model, better reflecting the sales data distribution, thus enhancing the pipeline's reliability.

\subsection{Visual Search-Based Models} \label{sec:sec_visual_search_models}
In~\cite{joppi2022pop}, the authors introduce a novel data-centric pipeline that generates exogenous observation data using web-sourced information, termed POtential Performance (POP).
Specifically, the main objective of the POP signal is to predict how a new product might have performed in the past, providing insight into its future sales potential.
This approach uses only a single product image and freely available web data (such as images) to create a robust predictive signal.

In particular, the pipeline uses web-based learning to gather fashion trends over time, making it independent of proprietary or costly datasets.
This approach allows for a free and scalable collection of fashion-related data, representing both fashionable and unfashionable items, which are used to train a binary classifier.
This work is grounded in Data-Centric Artificial Intelligence (DCAI)~\cite{mcauley2015image}, emphasizing the automation of the creation of high-quality training data.
Using images and textual data from the Web, the authors automate the process of generating a reliable exogenous signal for sales forecasting models.

The proposed pipeline operates by generating the POP signal from a single image of the new product, referred to as the probe image.

The pipeline starts by extracting textual tags associated with the product image.
These tags, which describe attributes such as category, color, and fabric, can be taken manually from the technical sheet or automatically generated. 

Secondly, the ``query expansion'' and ``image retrieval'' steps consist of retrieving and expanding the extracted tags into positive (fashionable) and negative (unfashionable) tags.
Using these tags, the system performs time-specific queries online, retrieving images from past periods, which correspond to the periods where these trends were prominent or fading.

As a third step, the retrieved images are used to train a robust binary classifier through confident learning.
This classifier learns to distinguish between fashionable and unfashionable items for each previous interval.
The confident learning approach prunes noisy or irrelevant images from positive and negative classes, ensuring that the model is trained on clean, high-quality data. 

After the confident learning, the last step of the pipeline aims to generate the POP signal.
The pruned positive images are projected into an embedding space using the learned classifier, and the same is done for the original probe image.
The similarity between the probe and these past images generates the POP signal, which reflects how well the probe would have aligned with the past fashion trends.
The resulting POP signal can be integrated into every \gls{nfppf} model as input data.

\section{Emerging Trends and Outlook} \label{cha:cha_trends}

A major difficulty in predicting fashion trends lies in the ever-changing nature of consumer tastes.
Fashion does not follow a regular cycle; trends continuously shift and change.
In addition, emerging technologies, cultural changes, and unforeseen events can change existing trends.

\begin{table*}[!t]
    \centering
    \caption{State-of-the-art methods for \gls{nfppf}.
    The table lists the modalities used by the various methods, along with their results in terms of MAE and WAPE on the VISUELLE dataset.
    \textbf{I} stands for \textit{``Image''}, \textbf{R} for \textit{``Release Date''}, \textbf{D} for \textit{``Description''} and \textbf{G} for \textit{``Google Trends''}.
    In \textbf{bold}, the best results.
    \underline{Underlined}, the second best.
    (*) indicates that the model uses description as a further modality retrieved from other image-to-text models.}
    \resizebox{\textwidth}{!}{\begin{tabular}{l|cccc|c|cc}
    \toprule
    \textbf{Method} & \textbf{\textsc{I}} & \textbf{\textsc{R}} & \textbf{\textsc{D}} & \textbf{\textsc{G}} & \textbf{Architecture} & \textbf{WAPE $\downarrow$} & \textbf{MAE $\downarrow$} \\
    
    \midrule
    Attribute k-NN~\cite{ekambaram2020attention}    & \xmark & \xmark & \cmark & \xmark & k-NN                 & 59.8 & 32.7\\
    Image k-NN~\cite{ekambaram2020attention}        & \cmark & \xmark & \xmark & \xmark & \gls{convnet} + k-NN & 62.2 & 34.0\\
    Attr+Image k-NN~\cite{ekambaram2020attention}   & \cmark & \xmark & \cmark & \xmark & \gls{convnet} + k-NN & 61.3 & 33.5\\
    GBoosting~\cite{friedman2001greedy}             & \xmark & \cmark & \xmark & \xmark & Grad. Boost.         & 64.1 & 35.0\\
    GBoosting+G~\cite{friedman2001greedy}           & \xmark & \xmark & \xmark & \cmark & Grad. Boost.         & 63.5 & 34.7\\
    Cat-MM-RNN~\cite{ekambaram2020attention}        & \xmark & \cmark & \cmark & \cmark & \gls{rnn}            & 63.3 & 34.4\\
    X-Att-RNN~\cite{ekambaram2020attention}         & \cmark & \cmark & \cmark & \xmark & \gls{rnn}            & 59.5 & 32.3\\
    MDiFF~\cite{avogaro2024mdiff}                   & \cmark & \cmark & \xmark & \xmark & \gls{ddpm}           & 54.7 & 30.1\\
    Dif4FF~\cite{avogaro2024dif4ff}                 & \cmark & \cmark & \xmark & \cmark & \gls{ddpm}           & 54.6 & 30.0\\
    GTM-Transformer~\cite{skenderi2024well}         & \cmark & \cmark & \cmark & \cmark & Transformer          & 55.2 & 30.2\\
    \gls{muquar}~\cite{papadopoulos2022multimodal}* & \cmark & \cmark & \xmark & \cmark & \gls{mlp}            & \underline{52.6} & \underline{28.7}\\
    POP~\cite{joppi2022pop}                         & \cmark & \cmark & \cmark & \cmark & Transformer          & \textbf{52.3}    & \textbf{28.6}\\
    \bottomrule
\end{tabular}}
    \label{tab:tab_sota_models}
\end{table*}

In~\cref{cha:cha_nfppf_taxonomy}, we reported a comparative analysis of all models available today to tackle \gls{nfppf}, and here, in~\cref{tab:tab_sota_models}, we summarize it, providing for each of them the type of modalities used.
The results are reported in terms of Mean Average Error (MAE) and Weighted Absolute Percentage Error (WAPE)~\cite{hyndman2008forecasting}.
Formally, they are defined as:
\begin{equation} \label{eq:eq_mae}
    \text{MAE}=\frac{\sum_{t=0}^T |y_t-\hat{y_t}|}{T}\;,
\end{equation}
\begin{equation} \label{eq:eq_wape}
    \text{WAPE}=\frac{\sum_{t=0}^T |y_t-\hat{y_t}|}{\sum_{t=0}^T y_t}\;,
\end{equation}
where $y$ represents the actual values of the time series, $\hat{y}$ represents the forecasted values, and $T$ represents the total number of observations in the time series.

Our research shows that models relying only on images and tabular data outperform models that leverage external sources of knowledge such as~\cite{joppi2022pop,skenderi2024well}.  
In fact, while Google Trends can indicate search popularity, it does not guarantee that the item searched matches the exact appearance and features of the target garment.
Supplementing Google Trends data with visual search results has improved the performance of forecasting models.
By comparing the target garment with similar items found through Google Images, it is possible to gain a more comprehensive understanding of its potential appeal and market fit.
The pipeline proposed by~\cite{joppi2022pop} has shown state-of-the-art results but is still suboptimal.

Although the proposed pipeline is very promising, the problem of image filtering is not completely solved, even through the introduction of confident learning.
Indeed, there are many cases where a Google Image search using a specific prompt may result in images not completely relevant to the image probe.
An intuitive example of this shortcoming may be a Google search for images related to a skirt.
The main resulting images will often be of complete outfits, including the upper body. 

By the time any network is used to process the image and produce an embedding, there is no certainty that the resulting features are related to the skirt.
They could be related to the person's somatic features, shoes, shirt, or background.
For this reason, possible future developments regarding the retrieval of web-sourced information should consider these issues.

\subsection{Human-Pose Integration}
The automatic analysis of the human pose represents a source of valuable features in a wide range of tasks and domains~\cite{sampieri2022pose,avogaro2023markerless,avogaro2024exploring}, and its integration into fashion and garment fitting models presents several possibilities to improve the accuracy of \gls{nfppf}.
One key development in this field is to use detailed pose and mesh analysis to enhance \gls{nfppf} models.
By incorporating the knowledge of human body poses into these models, we can significantly improve how garments are represented and how well they align with different body types, enhancing both the virtual fitting~\cite{li2023virtual} experience and the overall garment design process.

By locating key points on the human skeleton, such as shoulders, hips, knees, and elbows, it is possible to gain insight into how a particular piece of clothing, like a skirt or jacket, will fit and behave on a person during movement.
This information can then be integrated into the \gls{nfppf} models to predict how different fabrics, cuts, or garment designs will adapt to various types of bodies and poses.
Human pose analysis can also streamline image selection and filtering during fashion-related web searches.
By applying 2D pose estimation techniques, we can extract a person's skeletal structure from an image and focus only on certain areas of the body.
For example, if you are analyzing an image to extract information about a skirt, pose estimation allows you to focus only on the key points related to the lower body, such as the hips and knees.
This level of detail means that irrelevant parts of the image, such as the upper body or background, can be filtered out, making the search more effective and targeted.

Pose-based filtering becomes particularly valuable when dealing with large-scale image datasets, such as those retrieved from web searches.
For example, if you used an image probe to look for skirts, pose estimation could help focus on the correct portion of the images, filtering out other irrelevant content.
This selective approach to image analysis helps reduce noise in search results and provides more accurate visual data.
It ensures that only relevant garments are processed in the correct body areas, ultimately improving the relevance of the retrieved results.

When combined with high-performance image segmenters, pose analysis reaches an even greater level of accuracy.
Systems such as SAM (Segment Anything Model)~\cite{kirillov2023segment} allow for detailed segmentation based on user-defined key points or regions of interest.
For example, once the pose of a person in an image has been extracted, SAM can be conditioned to segment only the regions of the image that correspond to the garment of interest, such as a skirt or a shirt.
By overlapping the segmented area with the pose key points, hips, knees, or shoulders, it becomes possible to isolate only the garment, removing the surrounding body parts or irrelevant background.

This kind of integration enables the precise extraction of garments for further analysis.
In addition, negative key points (those unrelated to the garment) can be used to enhance the segmentation process.
For example, if you analyze a skirt, key points associated with the upper body, such as the shoulders or head, could be treated as negative inputs.
These negative points would help the system filter out unnecessary areas, ensuring that the segmented image focuses solely on the analyzed garment.
This method not only refines the image search but also allows for more accurate comparisons of garment features, such as fabric, design, or fit.

\subsection{Leveraging Foundation Models Knowledge}
To expand on how Large Language Models (LLMs)~\cite{yao2024survey} like ChatGPT~\cite{openai2024chatgpt} and image generation models such as Stable Diffusion~\cite{rombach2022high}  can be used to retrieve information, we must first consider their capabilities for synthesis and interpreting large volumes of data.
These foundation models are trained on vast datasets that contain text or images that reflect various societal trends, including fashion, cultural shifts, and evolving consumer preferences over time.

\textbf{\emph{Retrieving historical trends through LLMs and image models.}}
LLMs, such as ChatGPT or Gemini~\cite{team2023gemini}, are designed to generate coherent and contextually relevant text based on the vast amounts of information on which they have been trained.
When applied to historical data, they can retrieve detailed information about past trends, including fashion preferences, social behavior, and even the popularity of certain garments or accessories.
For example, by querying these models with a specific time frame or fashion era, users can retrieve insights into how popular a particular style of clothing was in the 1990s or how trends in streetwear evolved in the 2010s.
This ability to generate and interpret nuanced information can replace traditional web scraping methods or combing through archives for fashion analytics, offering a faster and more efficient approach.

Similarly, models such as Stable Diffusion or DALL-E~\cite{ramesh2021zero}, which are designed to generate images, can be instrumental in visually interpreting fashion trends.
Based on their training data, these models can generate images that align with popular trends from specific periods.
For example, by asking Stable Diffusion to generate images of street fashion from the 1980s, one can visually assess what clothing styles, color schemes, and accessories were prevalent during that era.
This type of image generation can offer insight into the visual culture of past decades, aiding in historical analysis of fashion, or even helping brands revive vintage styles.

\textbf{\emph{Replacing retrieval pipelines with generative data.}}
Considering that LLM models are trained on a massive corpus of data that reflects human behavior and trends, it becomes plausible to replace data retrieval pipelines with entirely \gls{ai}-generated data.
For example, in the original pipeline presented by~\cite{joppi2022pop}, in order to understand whether or not a garment could be fashionable, the authors tried to compare the target with elements derived from Google Images.
The images searched on the web could be replaced with images generated by image-based foundation models.
With their vast knowledge of fashion trends, consumer behavior, and cultural shifts, these models can provide insights that would traditionally require extensive manual research.
Not only does this streamline the entire data collection process, but it also ensures that the information retrieved is grounded in vast and comprehensive datasets that cover a wide temporal and geographical range.

Another key advantage of these models is their ability to break down data by time intervals, provided that they are trained with temporal indexes or databases that categorize information according to historical periods.
For example, an LLM trained on fashion articles from the 1950s to the present day could be queried to retrieve information specifically from the 1980s.
This time-segmented analysis can be crucial in fashion forecasting or historical research, as it allows users to pinpoint exact periods of interest and retrieve accurate information relevant to that time frame.

This capability becomes especially useful in the context of tools like POP~\cite{joppi2022pop} and GTM~\cite{skenderi2024well}, where analyzing data from the correct time intervals is crucial before releasing a new garment. 
These \gls{ai} models can break down trends and popular tastes in specific time windows, making it easier to predict how a garment might perform in the current market based on past data. 
This kind of temporal filtering could enhance the analysis's precision.

\section{Conclusion} \label{cha:conlusion}

The rapid evolution of fashion trends, driven by consumer preferences, cultural shifts, and technological advances, makes \gls{nfppf} a critical challenge (\cref{cha:cha_introduction}).
This survey provides a comprehensive review of existing approaches, highlighting the strengths and limitations of different methodologies in addressing the \gls{nfppf} problem.
As summarized in~\cref{fig:fig_prisma}, this paper presents a systematic review of \gls{nfppf} from January 2020 to December 2024 (\cref{cha:cha_selection_criteria}).
In~\cref{cha:cha_nfppf_taxonomy}, we have seen that traditional models that rely on structured tabular data are now being complemented by multimodal approaches that integrate visual data, real-time trends, and external market signals.
Finally,~\cref{cha:cha_trends} highlights the research areas and unsolved challenges that can lead to further improvements in the development of new \gls{nfppf} methodologies.

\section*{Acknowledgments}
This study was conducted within the MICS (Made in Italy – Circular and Sustainable) Extended Partnership and received funding from Next-Generation EU (Italian PNRR – M4 C2, Invest 1.3 – D.D. 1551.11-10-2022, PE00000004). CUP MICS D43C22003120001 - Cascade funding project CollaborICE.
Furthermore, this study was also carried out within the PNRR research activities of the consortium iNEST (Interconnected North-Est Innovation Ecosystem) funded by the European Union Next-GenerationEU (Piano Nazionale di Ripresa e Resilienza (PNRR) – Missione 4 Componente 2, Investimento 1.5 – D.D. 1058  23/06/2022, ECS\_00000043).
This manuscript reflects only the Authors’ views and opinions.
Neither the European Union nor the European Commission can be considered responsible for them.

\bibliography{sn-bibliography}
\end{document}